%% file: fibonacci.tex
 \documentclass[sigconf,nonacm]{acmart}
\usepackage{floatpag}
\floatpagestyle{empty} 

\copyrightyear{2022}
\acmYear{2022}
\setcopyright{rightsretained}
\acmConference[GECCO '22 Companion]{Genetic and EvolutionaryComputation Conference Companion}{July 9--13, 2022}{Boston, MA, USA}
\acmBooktitle{Genetic and Evolutionary Computation ConferenceCompanion (GECCO '22 Companion), July 9--13, 2022, Boston, MA, USA}
\acmDOI{10.1145/3520304.3528878}
\acmISBN{978-1-4503-9268-6/22/07}

\newlength{\halfwidth}
\newlength{\temp}
\newlength{\tempa}

\begin{document}

\title{Failed Disruption Propagation in Integer Genetic Programming}

\author{\href{http://www.cs.ucl.ac.uk/staff/W.Langdon}
{William B. Langdon}}
\email{W.Langdon@cs.ucl.ac.uk}
\affiliation{%
  \institution{Department of Computer Science, University College London}
  \streetaddress{Gower Street}
  \city{London}
  \state{}
  \country{UK}
  \postcode{WC1E 6BT, UK}
}

\begin{abstract}
We inject a random value into
the evaluation of highly evolved deep integer GP trees
9\,743\,720 times
and find
99.7\% of test outputs are unchanged.
Suggesting crossover and mutation's impact are dissipated
and seldom propagate outside the program.
Indeed only errors near the root node have impact
and disruption falls exponentially with depth
at between $e^{-{\rm depth}/3}$ and $e^{-{\rm depth}/5}$
for recursive Fibonacci GP trees,
allowing five to seven levels of nesting 
between the runtime perturbation
and an
optimal test oracle for it to detect most errors.
Information theory explains this locally flat fitness landscape 
is due to FDP\@.
Overflow is not important and instead,
integer GP,
like 
deep symbolic regression floating point GP 
and software in general,
is not fragile,
is robust,
is not chaotic and
suffers little from Lorenz' butterfly.
\end{abstract}


\begin{teaserfigure}
\centerline{%
\includegraphics[width=0.5\textwidth]{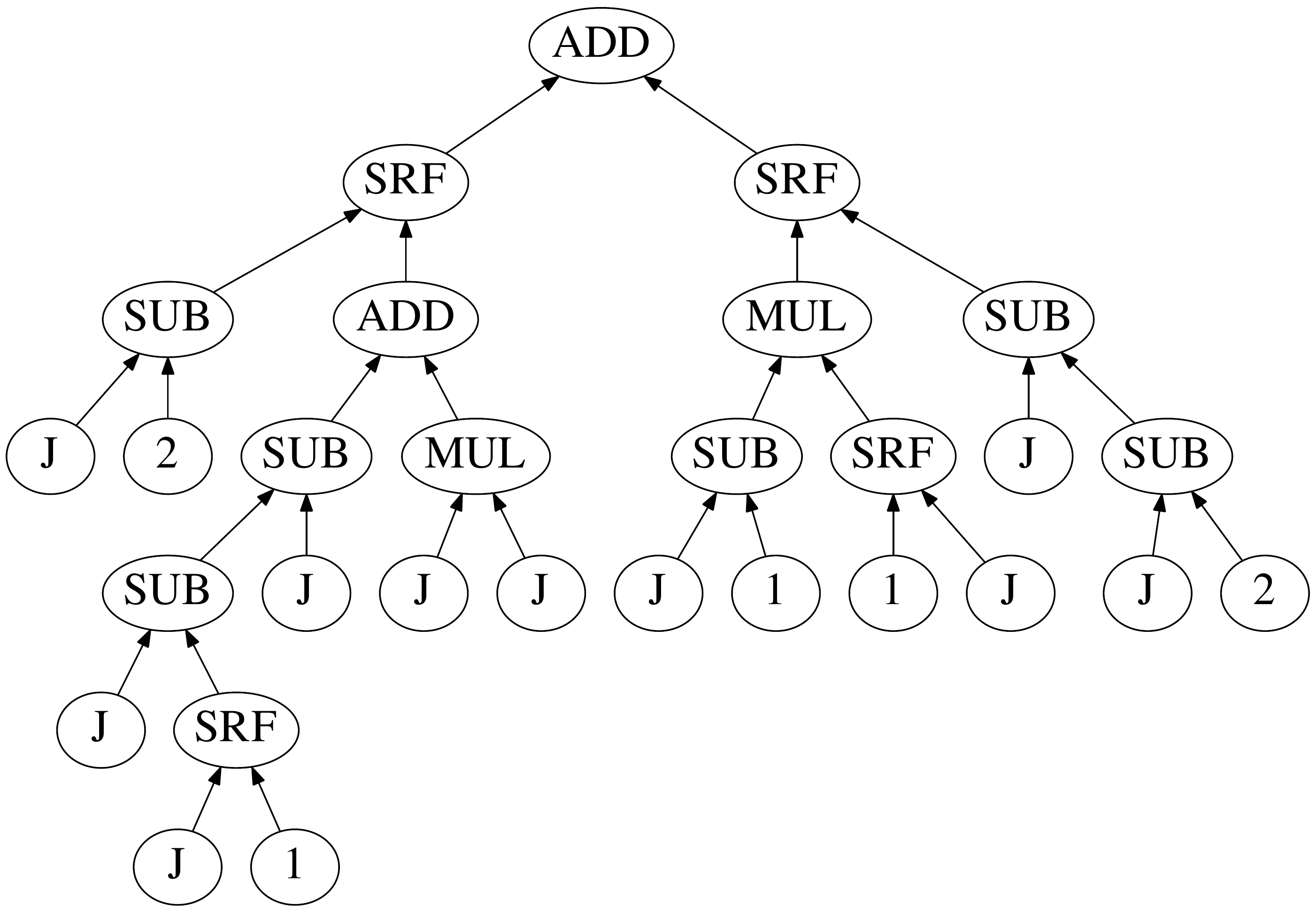}  
\includegraphics[width=0.5\textwidth]{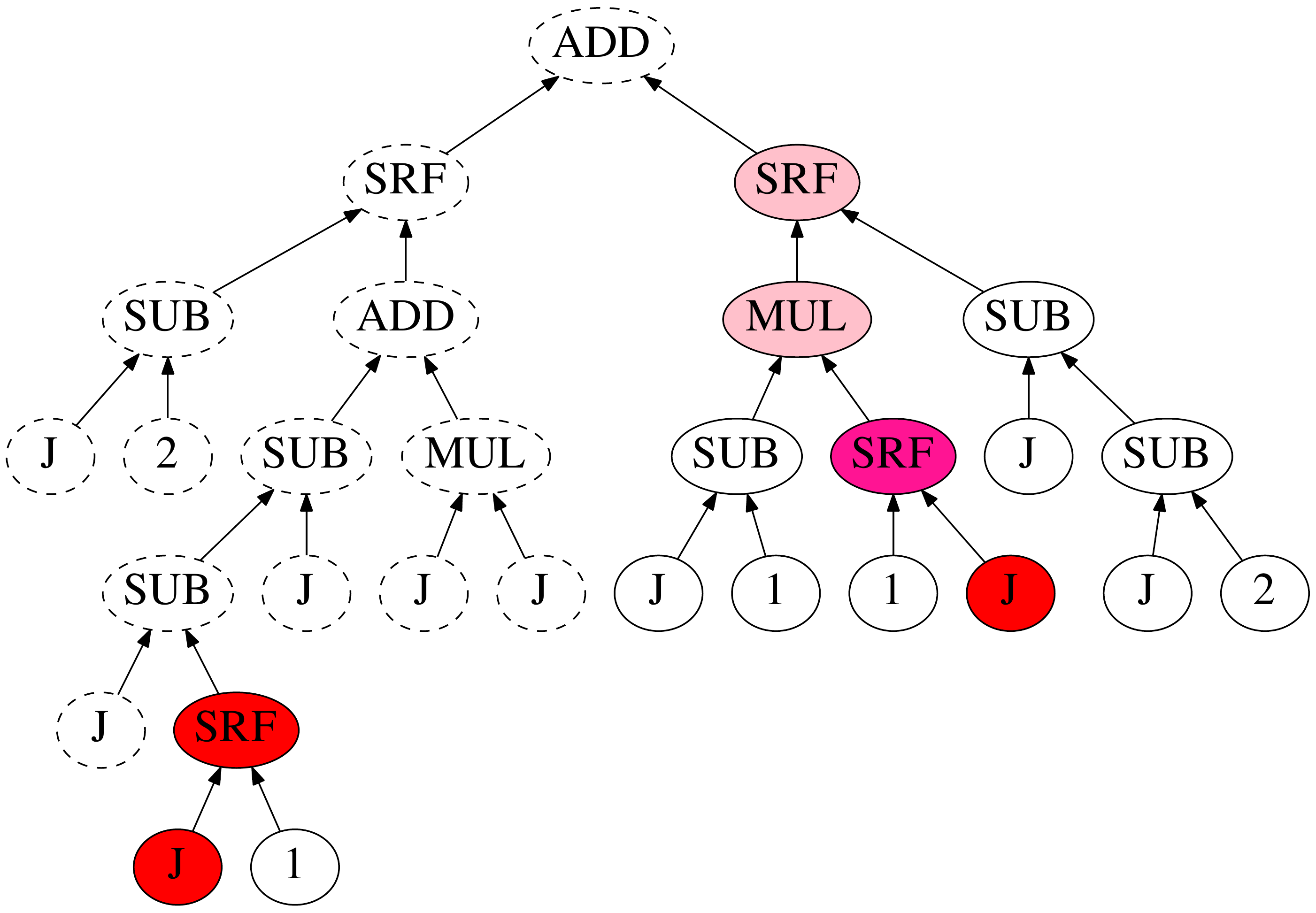} 
}
\vspace*{-2ex}
\caption{Left:
an evolved Fibonacci solution.
The arrows show the information flow from the leafs to the root node
(top oval).
The root gives the output of the whole program.
{\bf Right:}
Evaluation of the tree
showing two examples of run time disruption.
1)~When the bottom most J node (in red) is artificially perturbed by +1,
the disruption only reaches the calling SRF function (in red),
whose output does not change.
2)~Similarly, on most test cases,
disrupting the other red J node by +1,
means its calling SRF node's (dark pink) output does not change.
On test case J=0, the disruption propagates 3 levels to the pink SRF node.
On test case J=1, the disruption propagates as far as the pink MUL node.
(See page~\pageref{p.Danglot} and
Sections~\protect\ref{sec:+1} and~\protect\ref{sec:RANDINT}.)
}
\label{fig:fibonacci}
\end{teaserfigure}

\maketitle

4 page version to appear in
GECCO ’22 Companion, July 9–13, 2022 
ACM ISBN 978-1-4503-9268-6/22/07.\\
\href{https://doi.org/10.1145/3520304.3528878}
{https://doi.org/10.1145/3520304.3528878}

\nocite{Ibarra-Vazquez:2021:GECCO} 

\newpage
\section{Information Theory}

We can view computing as processing information
\cite{Renyi:diaryIT}.
Information in the program's inputs is transformed
as it passes through the program 
and emerges at its outputs.
In deterministic programs,
there is no information gain.
Indeed only in the special case that
the program is reversible
\cite{langdon:2003:normal,langdon:2007:gpem}
is the amount of information (number of Shannon bits)
in the output the same as that which went into the program.
In real programs, the information content is reduced.
That is computing destroys information.

Further individual operations inside a digital computer
may destroy information.
Excluding the special case of reversible functions,
all the functions from which a program is made
individually loose information.
For example, a 32~bit addition operation takes two inputs
and creates one output.
The inputs may each contain up to 32~bits of information 
(total 64~bits)
but its output can contain no more than 32~bits of information.
Storing data in memory allows the program to retain information,
but when that data is used, its information is liable to be
reduced or even lost
\cite{
Clark:2020:facetav}.
Computation maps multiple input patterns to the same output.
For example: \mbox{3 + 5} and \mbox{2 + 6}, 
each give the same value,~8.
If we follow the input data's path through an executing program,
e.g.\ we trace 3,5 in one run and 2,6 in another,
where the paths meet, e.g.\ at +~giving~8,
there is potential entropy~\cite{Renyi:diaryIT} loss.
Note, from the value~8 we cannot tell if we started with 3,5 or 2,6.

Much of genetic programming is concerned with evolving functions
without side effects.
For example in symbolic regression GP evolves
trees which take data from (32~bit) floating point inputs
and generates a (32~bit) floating point output at its root node.
The information compression is even more dramatic
in GP binary classification problems
where all the information in the tree's inputs is reduced to 
at most a single bit.

\begin{figure}
\centerline{%
\includegraphics[height=2.5in]{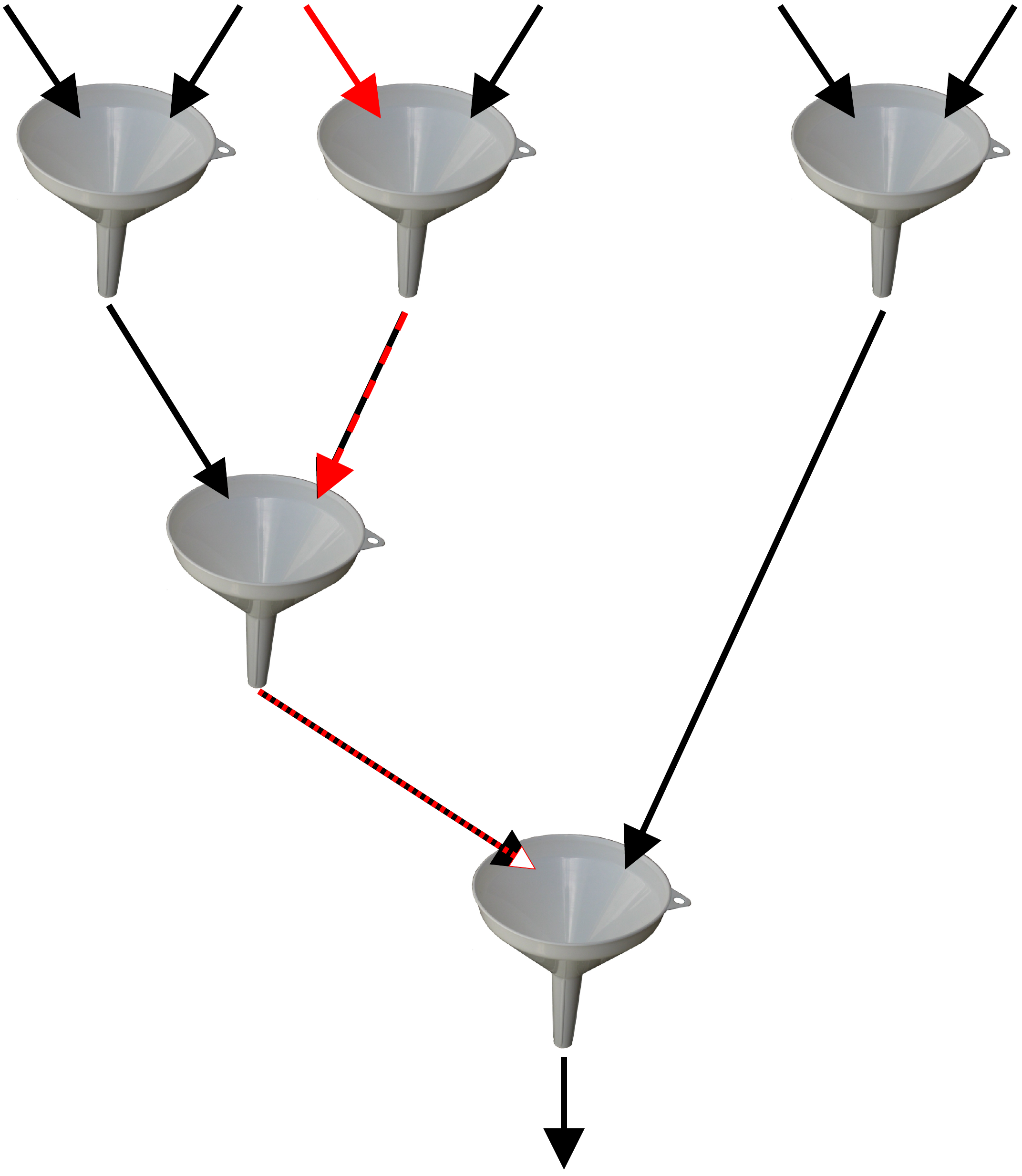} 
}
\vspace*{-2ex}
\caption{Each information funnel represents 
an irreversible function taking in total more
information than leaves it.
(Output at bottom.)
Disruption (in red) is progressively dissipated 
and may not change the output at all.
}
\label{fig:5funnels}
\end{figure}

We can view GP primitives as information funnels
\cite{langdon:2021:ieeeblog},
with wide mouths which take information from the function's inputs
and a narrower output,
corresponding to less information leaving the function.
In tree GP it is easy to see the whole tree as being made from
information funnels,
with one funnel per node in the tree,
see Figure~\ref{fig:5funnels}.
We can view crossover, mutation, and even runtime glitches,
as injecting a disruption into the GP tree.

If a deep GP tree is perturbed, the disruption has to propagate
from the crossover point, mutation or error, up the tree through many levels
to the root node before
it has any impact on the tree's fitness.
It is known in conventional programming \cite{%
kaetal:analysis,
Petke:2021:FSE-IVR}
that often disruptions fail to propagate.
We argue that this stems directly from information loss
and so is inherent in all computation, including GP\@.
We have shown that failed disruption propagation~(FDP)
can be common in deep
floating point expressions~\cite{langdon:2022:GECCO2}.
The mechanisms which cause FDP
can vary between programs.
For example in GP symbolic regression,
FDP is often associated with rounding errors~\cite{langdon:2022:GECCO2}.
To show an example which is independent of floating point rounding,
we will show (in {\bf Section~\ref{sec:experiments}})
that failed disruption propagation
also occurs in deep integer GP trees,
where calculations are performed exactly.
We find
small~(+1) and large (RANDINT, Section~\ref{sec:RANDINT}) run time disruptions
rapidly dissipate in similar ways.
{\bf Section~\ref{sec:fibonacci}} describes the integer
Fibonacci Problem we will use
and {\bf Section~\ref{sec:discuss}} 
discusses our results and their implications
for GP and software more widely.

\subsection{Disrupting Integer Arithmetic}

Much of Koza's first GP book~\cite{koza:book}
is concerned with either floating point or Boolean expressions.
However Koza also introduces the problem of inducing 
an integer tree which recursively generates
the Fibonacci sequence of positive integers. 
Therefore we shall use the Fibonacci Problem and demonstrate
in deep GP trees failed disruption propagation can also be
common in exact arithmetic.

We use GP to evolve large GP trees
by simply running to 1000 generations
rather than stopping at the first solution.
We then re-evaluate the whole tree on all the training cases
having first inserted a fixed run time perturbation at a given
location.
We step though every location in
this large highly evolved fit tree
and keep a record of which perturbations do or do not
cause a change in evaluation at the root node
and on which test case.

\label{p.Danglot}
Like Danglot et~al.~\cite{danglot:hal-01378523},
the first perturbation is simply to add~1 to the evaluation
(on each test case)
at the chosen point in the tree.
This can be thought of as the minimum perturbation.
We also repeat the experiment but instead of
making a small change we simply totally replace
the original evaluation by a randomly chosen 32~bit signed
integer value.

\section{Koza's Fibonacci Benchmark}
\label{sec:fibonacci}

The Fibonacci sequence
1,
1,
2,
3,
5,
8...
is an infinite sequence of positive integers,
where the next one is given by summing the two previous 
items in the list.
For training we take the first 20 members of the sequence,
i.e.\ from the 0$^{\rm th}$~(1) to the 19$^{\rm th}$~(6765).
See Table~\ref{gp.details}.
The primitives Koza~\cite{koza:book} uses are
the four small integers 0, 1, 2 and~3,
the sequence index J,
the three integer arithmetic operations: 
addition, subtraction and multiplication.
In addition, to support recursion,
there is a special function SRF which also takes two arguments.
SRF allows access to values
calculated by the GP tree on {\em earlier} test cases.
SRF's first argument is the number of the test case.
SRF's second argument is a default value,
to be used if the first argument is invalid.
(For simplicity all arguments, 
including where we have multiplication by zero, are always evaluated.)
As an example \mbox{(SRF 1 0)} will evaluate to~0 
on the first two (J=0 and J=1) test cases,
and will evaluate to the evolved individual's
answer for test case~J=1 on later tests.
Test cases are always run in order, starting at J=0.
Notice, in Section~\ref{sec:experiments},
where tree evaluations are deliberately disrupted,
the disruption is applied per test case
and therefore SRF can access the earlier unchanged
tree evaluations.

\begin{table}
\setlength{\temp}{\halfwidth}
\settowidth{\tempa}{GP parameters: }
\addtolength{\temp}{-\tempa}
\addtolength{\temp}{-2\tabcolsep}
\caption{\label{gp.details}
GP to create deep fit Fibonacci trees
for 
\\
Failed Disruption Propagation (FDP) experiments
}
\vspace*{-2ex}
\begin{tabular}{@{}lp{\temp}@{}}\hline
Terminal set: \rule[1ex]{0pt}{6pt} & J, 0, 1, 2, 3\\
Function set:                      & ADD SUB MUL SRF \\
Fitness cases:                     & First 20 members of the Fibonacci sequence.
\\
Selection:  & Fitness =
$\sum_{\rm J=0}^{19}|{\rm GP(J)}-{\rm Fibonacci\,}_{\rm J}|$.
\ \
I.e.\ the
sum of the absolute error between GP's
answer and the value of the J$^{\rm th}$ member of the Fibonacci sequence.
Tournament size 7.
\\ 
Population: & 
Panmictic, non-elitist, generational.
\\ 
GP parameters: &
Initial population of 50\,000 trees created by ramped half and half
\cite{koza:book} with depth between 2 and~6.
100\%~unbiased subtree crossover.
1000~generations.
No size or depth limit.
\\\hline
\end{tabular}
\end{table}

\subsection{Background}

Perhaps because, by early GP standards,
the Fibonacci Problem needs a large population,
see Figure~\ref{fig:tune},
it has been little used in GP\@.
However 17 years after Koza's book was published,
At EuroGP 2009
Harding et~al.\ gave a nice summary \cite{Harding:2009:eurogp}.
Their survey includes
\cite{Huelsbergen:1997:lrsemlp}, 
\cite{nishiguchi:1998:erpmnGP},  
\cite{eurogp06:AgapitosLucas},  
\cite{1277290},                 
and
\cite{1277165}, 
all of these used approaches to the Fibonacci Problem 
which differ from Koza's~\cite{koza:book}.
Except Kouchakpour~\cite{Kouchakpour:thesis},
publications on inducing the Fibonacci sequence since EuroGP 2009
have all looked at non-standard GP\@.
They include
Castle's 2012 PhD thesis~\cite{Castle12}
which used it as one of half a dozen benchmarks to
compare Montana's strongly typed GP~\cite{montana:stgpEC}
with Castle's own Strongly Formed GP 
and 
with other higher level imperative primitive sets.
Also in 2012
Bryson and Ofria~\cite{10.1162/978-0-262-31050-5-ch003} 
used Avida to evolve solutions.
Whilst Atkinson's 2019 PhD thesis~\cite{Atkinson:thesis} 
looked at solution by evolving graphs. 
In 2018 Krauss et al.~\cite{Krauss:2018:GI5} 
used it with a very different AST representation 
in their genetic improvement \cite{Langdon:2012:mendel,%
Petke:2019:GI7} experiments.

Other work includes:
Gruau et al.~\cite{DBLP:journals/tcs/GruauRW95} who used a
Fibonacci program as an example to show coding Pascal programs
as neural networks.
Teller's 1998 PhD thesis~\cite{AstroTeller:thesis}
which includes it as a simple example of neural programming.
Yu who showed that programs generating the Fibonacci sequence can be evolved using higher-order functions~\cite{yu:2005:recurse}.
It has also been used to demonstrate Spector's Push~\cite{1068292} 
and by Binard~\cite{1274004} to demonstrate System~F\@. 
Finally Kouchakpour's 2008 PhD thesis
returned to Koza's setting of the Fibonacci Problem~\cite{Kouchakpour:thesis}.

\section{Experiments}
\label{sec:experiments}
\subsection{Tuning for Tournament Selection}

\begin{figure}
\centerline{\includegraphics{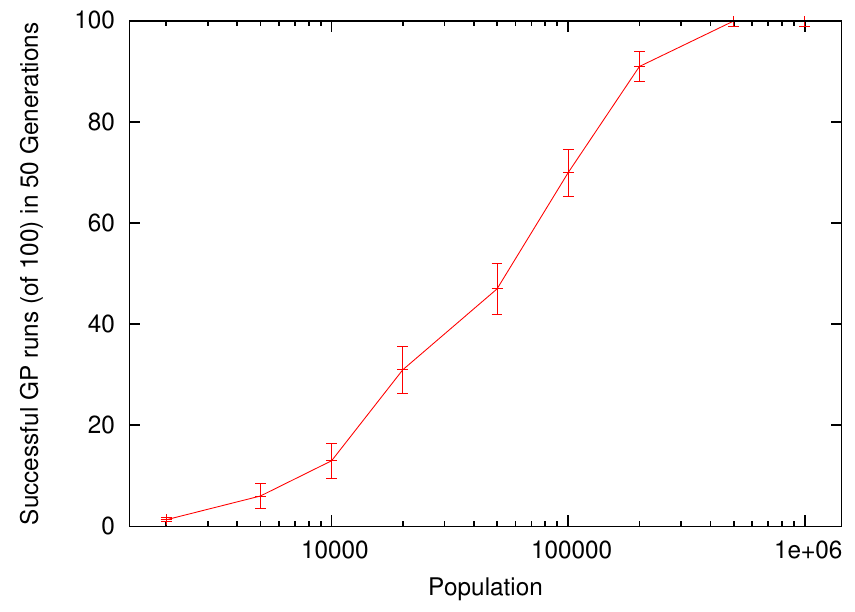}} 
\vspace*{-2ex}
\caption{Number of successful Fibonacci runs.
\label{fig:tune}
}
\end{figure}

We did a series of tuning GP runs 
to choose the population size
(Table~\ref{gp.details}).
From Figure~\ref{fig:tune} we can see that the chance of success
with tournament size~7,
separate, non-overlapping, generations 
(i.e.~with complete replacement)
and a population of 2000 trees is only 1.3\%.
So we chose a population of 50\,000,
where about 47\% of runs find programs
which pass all 20 Fibonacci fitness tests by generation 50.

\subsection{Ten Extended Runs to 1000 Generations}

\begin{figure}
\centerline{\includegraphics{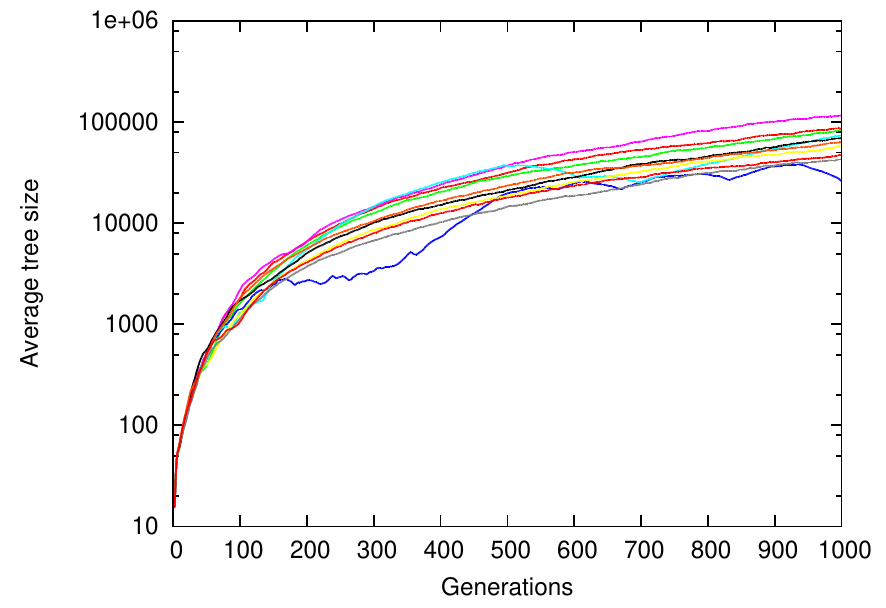}} 
\vspace*{-2ex}
\caption{Mean size of trees 
in ten extended GP Fibonacci runs
(population 50\,000).
Note log scale.
\label{fig:size}
}
\end{figure}

\begin{figure}
\centerline{\includegraphics{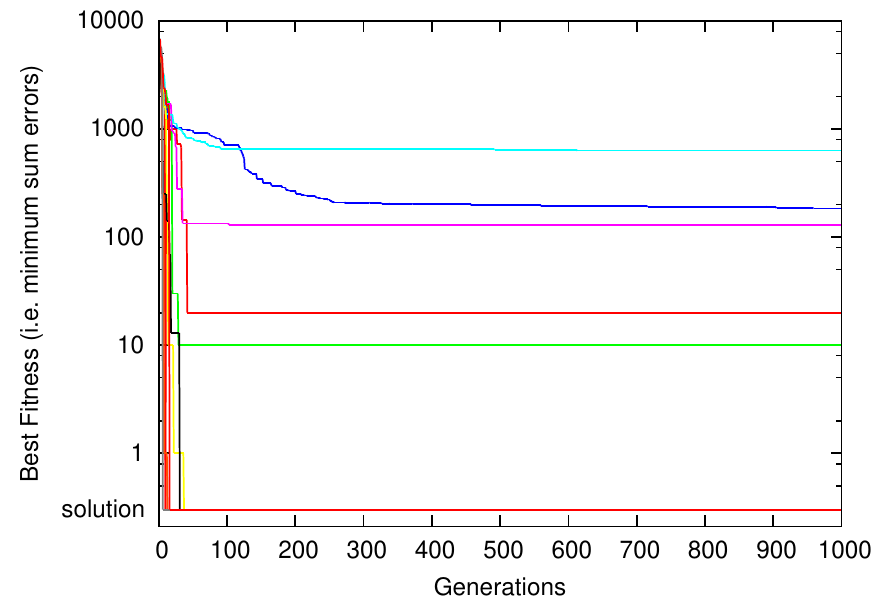}} 
\vspace*{-2ex}
\caption{Sum of training $|$error$|$
in ten extended GP \mbox{Fibonacci} runs
(population 50\,000).
Note log scale.
\label{fig:fit}
}
\end{figure}

\begin{figure}
\centerline{\includegraphics{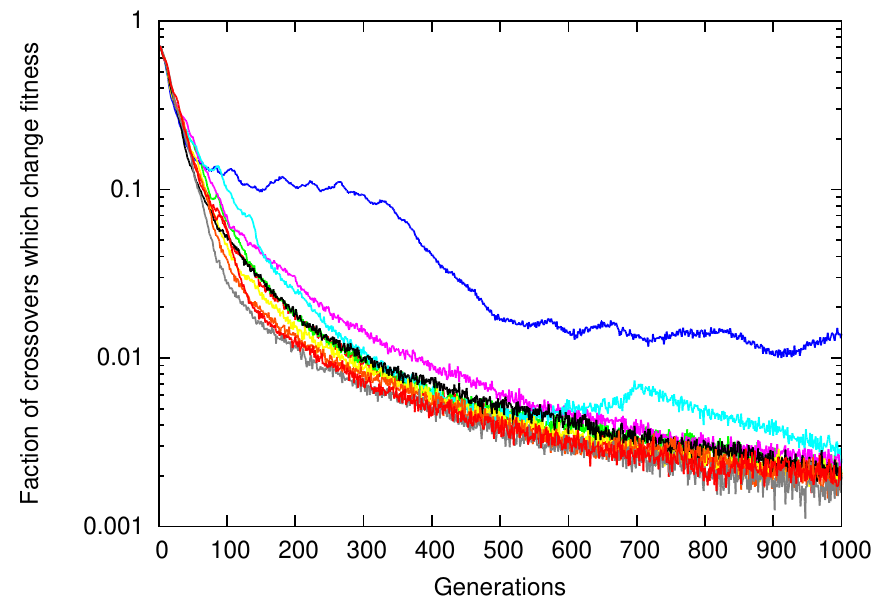}} 
\vspace*{-2ex}
\caption{Mean fraction of children with fitness different 
from their first (i.e.~root donating) parent
in ten extended GP \mbox{Fibonacci} runs
(population 50\,000).
Note log scale.
\label{fig:diff2}
}
\end{figure}

\begin{figure}
\centerline{\includegraphics{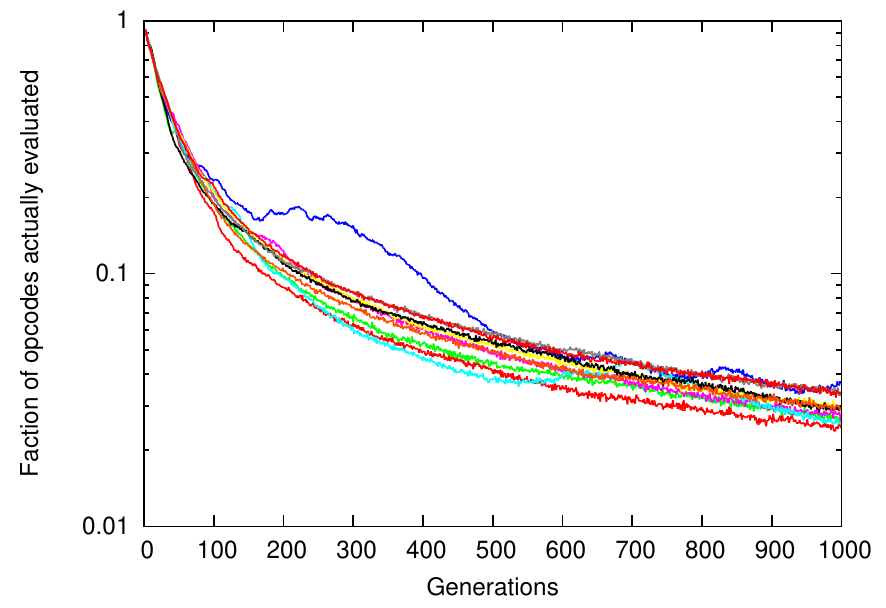}} 
\vspace*{-2ex}
\caption{Effectiveness of incremental evaluation
at exploiting FDP and convergence to reduce
the volume of GP opcodes that must be evaluated to calculate fitness.
(Ten extended GP Fibonacci runs with
population 50\,000).
By generation 1000, only 2.9\% of opcodes are evaluated.
Note log scale.
\label{fig:evals}
}
\end{figure}

\begin{figure}
\centerline{\includegraphics{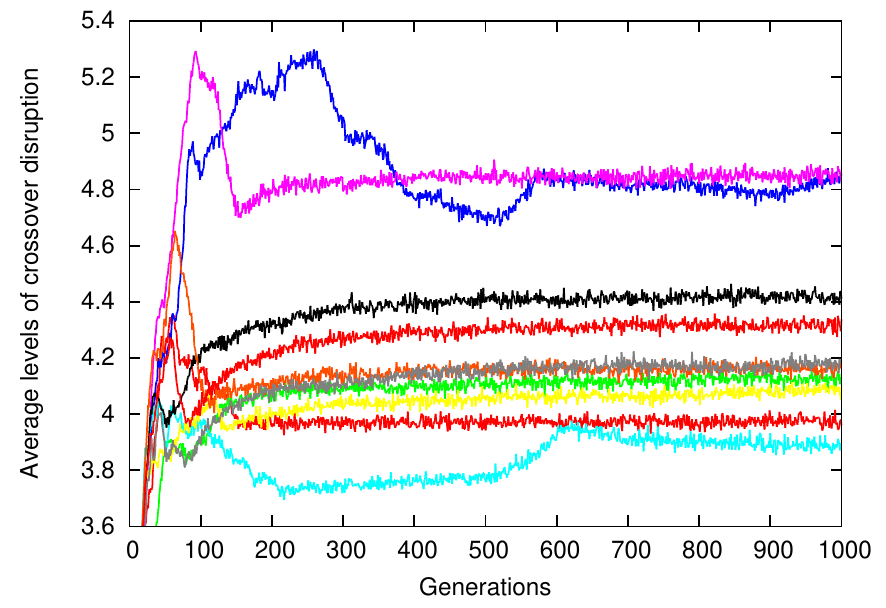}} 
\vspace*{-2ex}
\caption{Mean number of functions evaluated above crossover point 
per test case
in ten extended GP Fibonacci runs
using incremental evaluation~\protect\cite{langdon:2021:EuroGP}
(population 50\,000).
Note linear scale.
\label{fig:nups}
}
\end{figure}

In order to get deep fit Fibonacci trees
to try our disruption experiments on,
we ran our GP with a population of 50\,000
for 1000 generations.
(We used the same parameters, Table~\ref{gp.details}.)
Of course without size or depth limits,
the GP bloats~\cite{Langdon:1997:bloatWSC2}
(see Figures~\ref{fig:size} to~\ref{fig:nups}).
At the end of each run a Fibonacci tree was selected
for perturbation.
The runs are summarised in Table~\ref{tab:nups_summary}.
(Column~3 gives the expected depth for a random
binary tree of a given size, column~1,
whilst column~4 gives the standard deviation
\cite{flajlet:1982:ahbt}.)

\begin{figure}
\centerline{\includegraphics{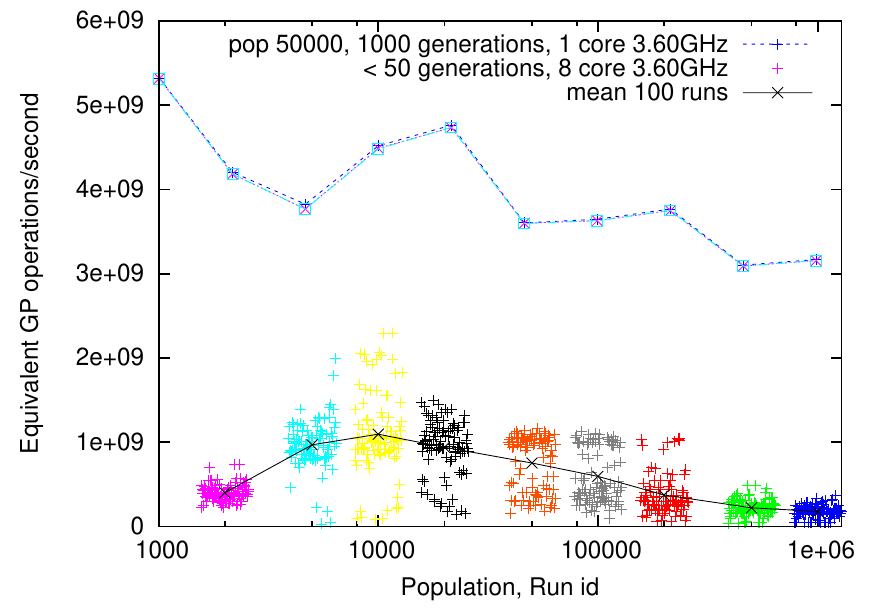}} 
\vspace*{-2ex}
\caption{GP speed.
{\em Lower}: 100 runs 
for each population size.
(For visualisation small x-direction noise added.)
{\em Top}: in ten 1000 generations runs
GP tends to converge
\cite{langdon:GPEM:gpconv}
making the speed up of incremental evaluation~%
\protect\cite{langdon:2021:EuroGP}
more effective.
Note top (single core) data are spread out horizontally for visualisation
only.
\label{fig:gpops}
}
\end{figure}

To reduce run time we used our~\cite{langdon:2021:EuroGP}
incremental
and ``fitness first'' \cite{Langdon:2021:GPTP} 
evaluation.
Of course this does not effect the course of evolution
but reduces the number of opcodes evaluated by between
27 and 41 fold.
(See Figure~\ref{fig:evals}.
Figure~\ref{fig:gpops}
gives the absolute speed in terms of GP operations/second,
GPops \cite{langdon:2008:SC}.)

\begin{table}
\caption{\label{tab:nups_summary}
Ten Deep Fit GP \mbox{Fibonacci} Trees}
\vspace*{-2ex}
\begin{tabular}{@{}rrr@{ (}c@{)}rr@{ \% }rr@{ \% }r@{}}
\multicolumn{1}{c}{Size} &
\multicolumn{2}{c}{Depth} &
\multicolumn{2}{r}{Fitness} & 
\multicolumn{4}{c@{}}{Output disruption on any test}  \\
& &
\multicolumn{1}{c}{
\cite{langdon:2000:quad}} & 
\multicolumn{2}{r}{sum $|$error$|$} & 
\multicolumn{2}{c}{+1} &
\multicolumn{2}{c@{}}{RANDINT} 
\\\hline
\input{graph/nups_summary2}\\
\end{tabular}
\end{table}

\subsection{Fraction of +1 Disruptive Perturbations}
\label{sec:+1}

In almost all locations in the ten deep highly evolved
GP Fibonacci trees,
the +1 disruption at run time
fails to reach the root node
on all twenty training cases.
The right hand side of Table~\ref{tab:nups_summary}
gives the fraction~(\%, column~6) which disrupt fitness on any
of the twenty test cases.
Even for the three shallowest trees more than 90\% of
the perturbations fail to propagate to the program's output
on any test case.
Therefore the fitness would be identical despite the runtime change.
In most cases the fraction of FDP on any of the fitness cases
is well in excess of 99\%
(max 99.968\% for run~9).

\begin{figure}
\centerline{\includegraphics{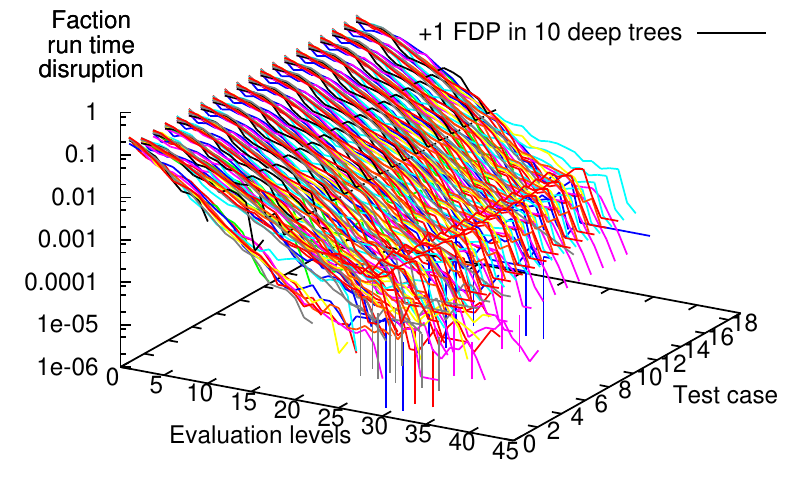}} 
\vspace*{-2ex}
\caption{How far +1 disruption travels up tree
for each training case (0--19).
Small fraction which reach the root node on any test
are given in Table~\protect\ref{tab:nups_summary}
and not plotted here.
Table~\protect\ref{tab:nups_summary} also gives median slope of these plots.
Note log scale.
\label{fig:fdp}
}
\centerline{\includegraphics{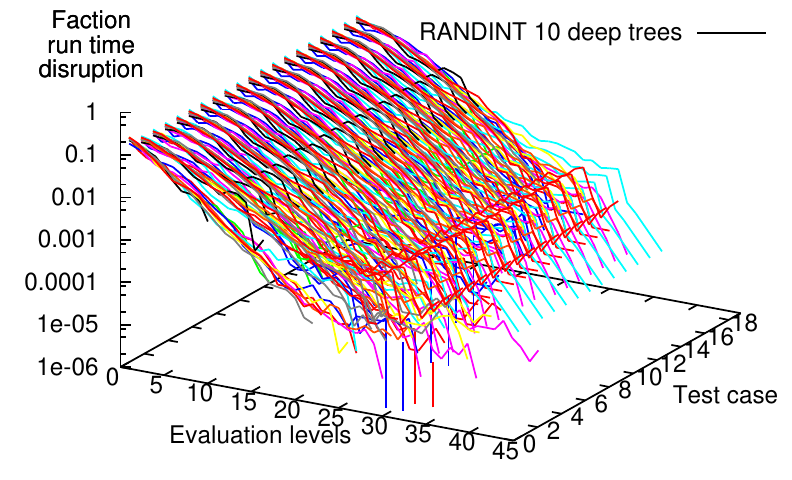}} 
\vspace*{-2ex}
\caption{How far RANDINT disruption travels up tree
for each training case (0--19).
Small fraction which reach the root node on any test
are given in Table~\protect\ref{tab:nups_summary}
and not plotted here.
Table~\protect\ref{tab:nups_summary} also gives median slope.
Note log scale.
\label{fig:fdp_RANDINT}
}
\end{figure}

In floating point symbolic regression~\cite{langdon:2022:GECCO2}
failed disruption propagation
depends on the magnitude of the test case
(with values near zero being suppressed most easily).
In contrast, 
Figure~\ref{fig:fdp} shows
in the Fibonacci Problem
almost the same behaviour for all test cases.
This suggests Fibonacci test cases are not independent.
Which in turn hints at potential runtime saving by sub-sampling test cases.
Only the zeroth test case (J=0) is slightly different,
and we see disruption dying away even faster
than in the other nineteen test cases.
Notice in Figure~\ref{fig:fdp}, FDP behaves similarly
across ten very different trees.
Each shows, for each test case,
a similar exponential decrease in the number of functions
the +1 disruption propagates through before becoming totally lost.
Column~7 in
Table~\ref{tab:nups_summary}
gives the exponent for the decrease with depth
averaged (median) across all twenty test cases.
The values lies between -0.33 and~-0.20,
meaning on average
between 14 and 23 nested functions will reduce
disruption by 100 fold.

\newpage
\subsection{Location of +1 Disruptive Perturbations}

\begin{figure}
\centerline{\includegraphics{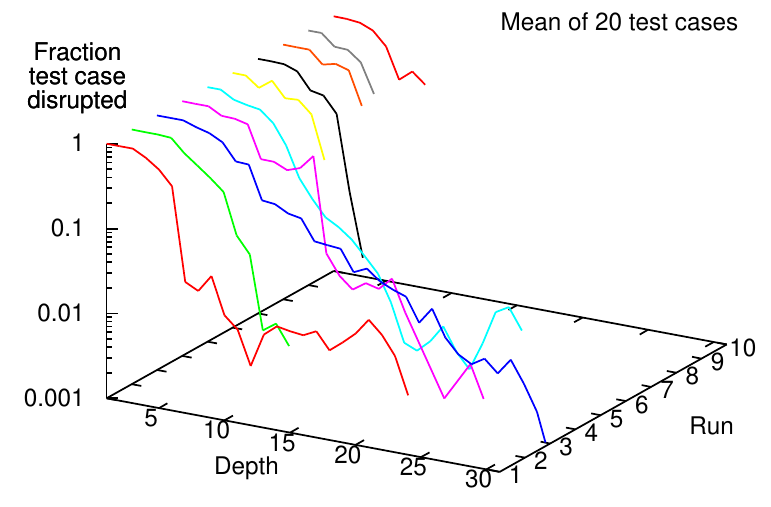}} 
\vspace*{-2ex}
\caption{Location of +1 disruptions which affect fitness
on any training case.
(See also left side of
Figures~\protect\ref{fig:lattice1} and~\protect\ref{fig:lattice2}.)
Note log scale.
\label{fig:depth}
}
\centerline{\includegraphics{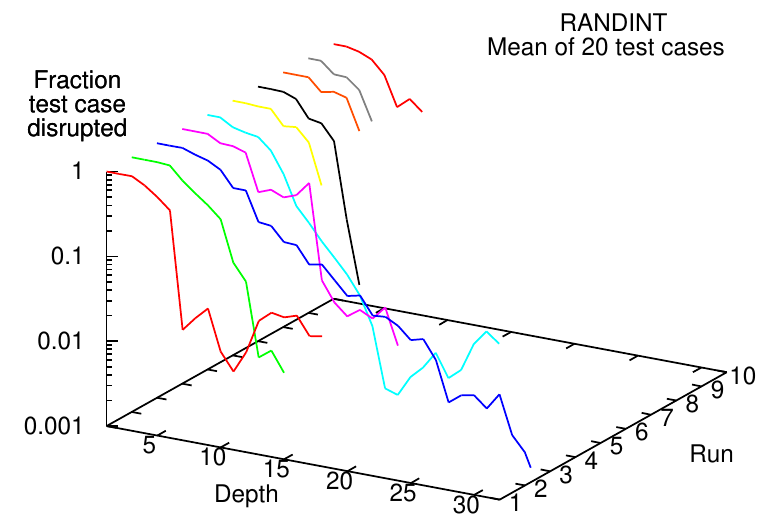}} 
\vspace*{-2ex}
\caption{Location of RANDINT disruptions which affect fitness
on any training case.
(See also right side of
Figures~\protect\ref{fig:lattice1} and~\protect\ref{fig:lattice2}.)
Note log scale.
\label{fig:depth_RANDINT}
}
\end{figure}

The small fraction of +1 disruptions which reach the root node,
on any test case,
and so do change fitness, 
lie close to the root node itself.
I.e, the disruption travels only a short distance through the code.
Depending on run,
most lie within 4--9 levels of the root node. 
For most runs (see Figure~\ref{fig:depth})
more than 99\% of these disrupted evaluations 
start within 8 nested function calls of the root.
Runs~1, 3 and~5 have long tails,
so that more than 1\% of disrupted evaluations start more than 20 levels deep.
Even so in these runs no disruption traverses more than 31 levels 
and still impacts fitness.

\subsection{Integer Overflow}
In the +1 FDP experiments,
except for one subtraction,
only integer multiplication leads to overflowing 32~bits.
The fraction of multiplication (MUL) node
outputs being truncated to 32~bits
varies considerably between runs.
In most cases there is none or very little ($<$1\%),
in others between 5\% and~23\% of disrupted multiplications overflow.
In all cases, +1 disruption which causes 32~bit integer overflow
is stopped by the usual mechanisms
(see next section)
and does not reach the program's output.

\subsection{Fibonacci Mechanisms for\\Failed Disruption Propagation}
\label{sec:mechanisms}

Although we argue that information theory suggests that
failed disruption propagation is universal,
we can see specific mechanisms for FDP in the Fibonacci Problem.
For simplicity let us just consider cases where the disruption
on all 20 test cases stops at the same point.
20\% of such +1 disruptions stop on a multiply by zero.
That is,
one argument of multiply is zero and so disruption to the
other fails to propagate past the multiply,
since its output is zero regardless of the disruption.
All the others stop on a SRF function.
Of these most (98.7\%)
are because the SRF node returned its default value
as it did before the disruption.
This may be because the SRF function's first argument 
(the index) was already invalid and so caused
SRF to return its default value (the 2$^{\rm nd}$ argument).
And if disrupting the index value still leaves it invalid,
the SRF will continue to return its unchanged default value.
Meaning the disruption stops at the SRF node.

\subsection{RANDINT Disruption}
\label{sec:RANDINT}
We repeated the above experiments 
replacing the small perturbation, 
which simply added one to the evaluation of each point
in each large GP tree for each test case,
by replacing the existing evaluation by
a random integer value.
The value was chosen uniformly at random from all $2^{32}$ possible 
signed integer values.
Then as before we trace how far the large disruption
propagates through the evolved tree.

Starting with Table~\ref{tab:nups_summary}
we see that the large RANDINT disruption behaves very similarly
to the +1 disruption.
Comparing the last four columns of Table~\ref{tab:nups_summary},
we see almost all large disruptions fail to reach the
root node and so make no difference to the program's
output or fitness.
And further the exponential rate 
(between $-1/3$ and $-1/5$)
with which the 
average 
disruption dies away with distance from the root node,
although different between runs,
is very similar for +1 and RANDINT\@.
Comparing Figures~\ref{fig:fdp} and~\ref{fig:fdp_RANDINT}
again shows little difference between
the distance traveled through the evolved code
by the large and small disruptions.
Whilst Figures~\ref{fig:depth} and~\ref{fig:depth_RANDINT}
show the small fraction of large and small disruptions which are able
to reach the program's output are
similarly clustered close to the output itself
(the root node).
Figures~\ref{fig:lattice1} and~\ref{fig:lattice2},
next section,
also show that +1 and RANDINT have similar patterns of
failed disruption propagation.

When disrupting by replacing an evaluation by a large value
(rather than making a small change),
both addition and subtraction may subsequently overflow
32~bits.
However multiplication continues to be by far the most likely
operation to cause overflow.
Again there is variation between runs
with between 30\% and 65\%
of disruptions leading to overflow.
Of the small fraction of RANDINT disruptions which reach the root
(column~8 in Table~\ref{tab:nups_summary}),
between 1\% and~38\% are affected by overflow.

\subsection{Disruptable code lies near the output}
Figures~\ref{fig:lattice1} and~\ref{fig:lattice2}
compare the impact of the +1 and random replacement perturbations.
(To reduce clutter, SRF functions are shown with =.)
The plots are each in five rows of horizontal pairs.
In each row the same program is shown.
With the impact of increasing each evaluation by +1
shown on the left
and the impact of replacement with a random 32~bit value
(RANDINT)
on the right.
The plots show the binary trees using Daida's circular lattice
\cite{daida:2005:GPEM}.
This can be thought of as viewing the binary tree from the top
looking down on the root node (in the center)
with each side subtree spread out around it.
The colours indicate how many times the program's answer
is changed when evaluation at that point in the program is disrupted
on each test case.

\setlength{\temp}{-52pt} 
\begin{figure*}
\vspace*{-64pt}
\centerline{
\includegraphics{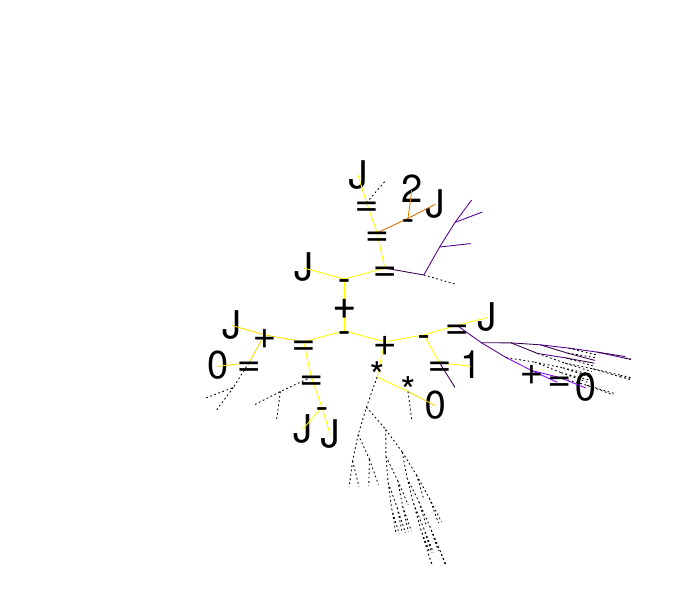}
\includegraphics{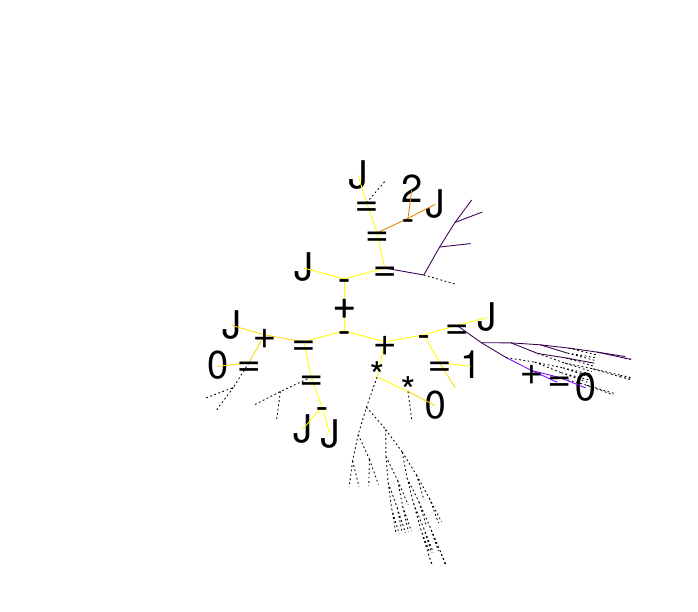}}
\vspace*{\temp}
\centerline{
\includegraphics{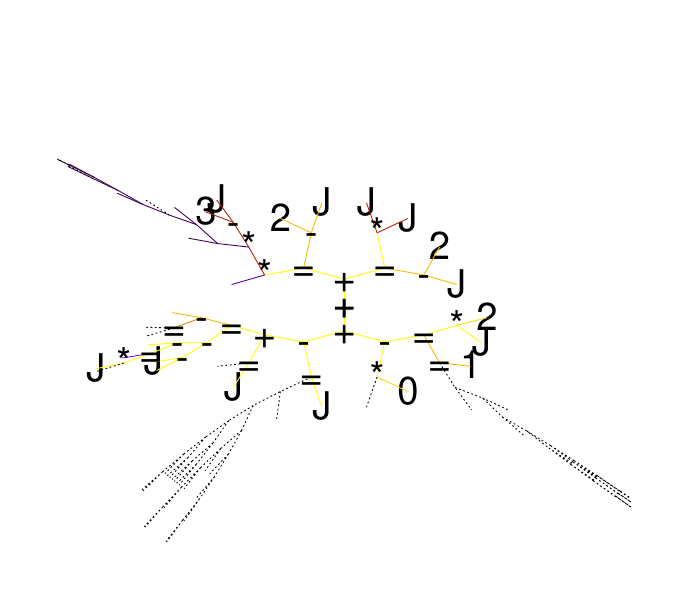}
\includegraphics{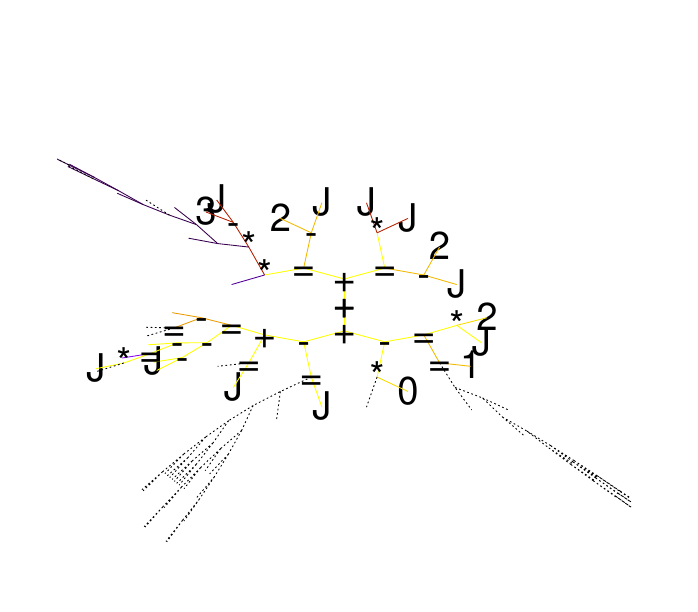}}
\vspace*{\temp}
\centerline{
\includegraphics{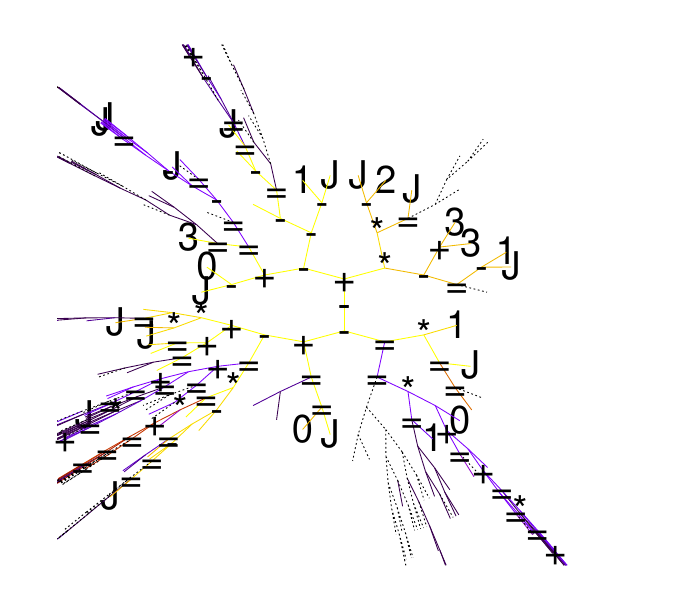}
\includegraphics{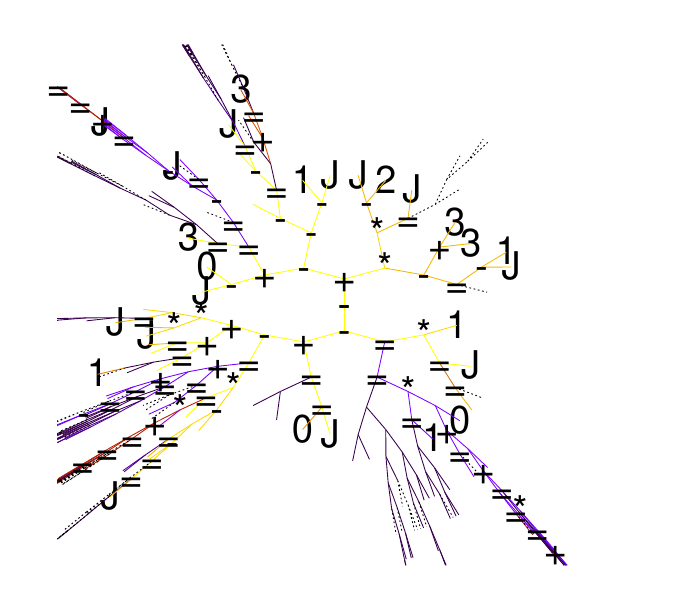}}
\vspace*{\temp}
\centerline{
\includegraphics{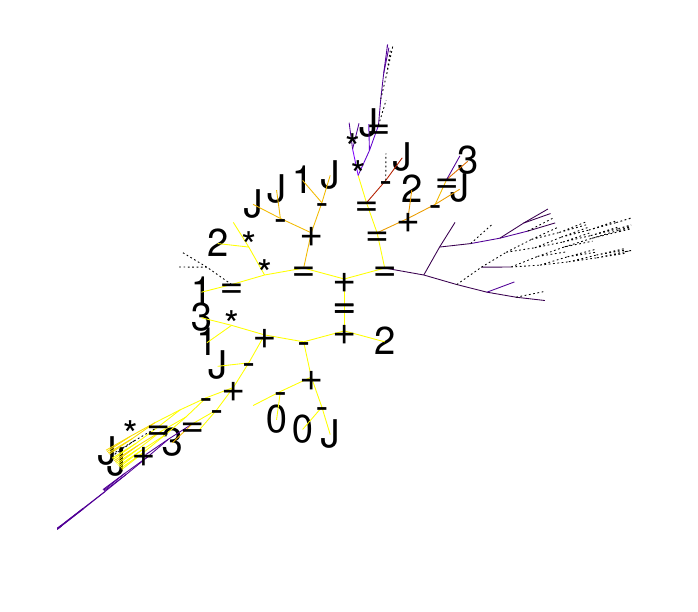}
\includegraphics{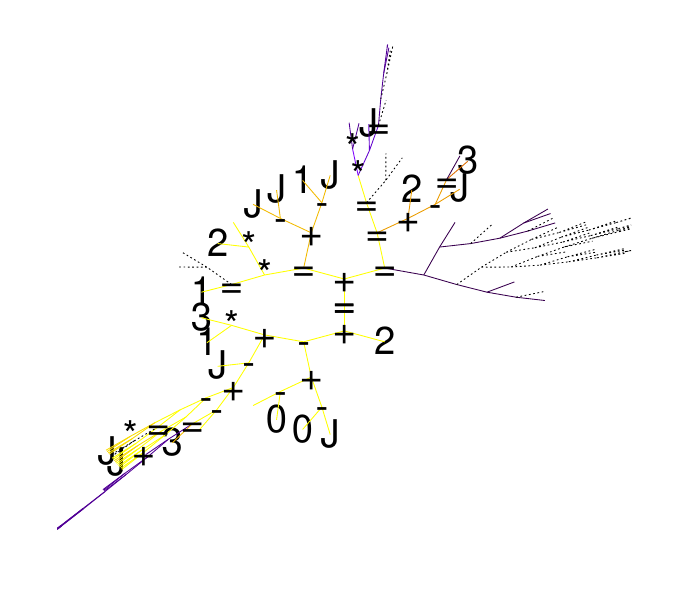}}
\vspace*{\temp}
\centerline{
\includegraphics{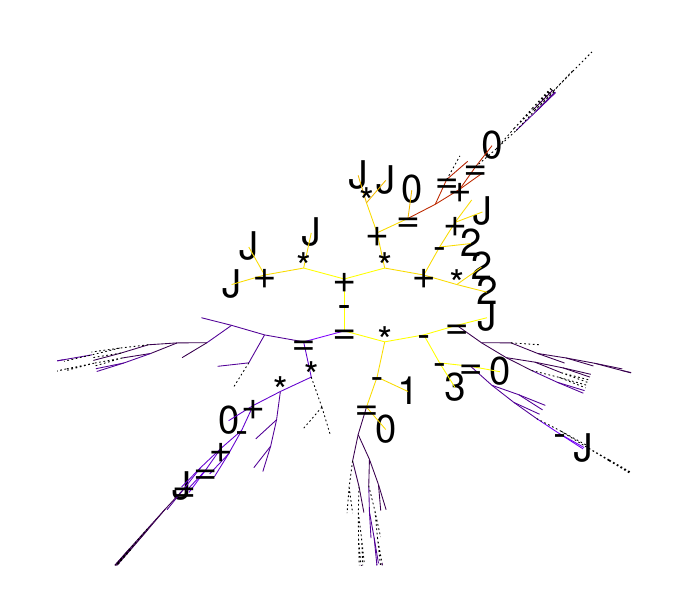}
\includegraphics{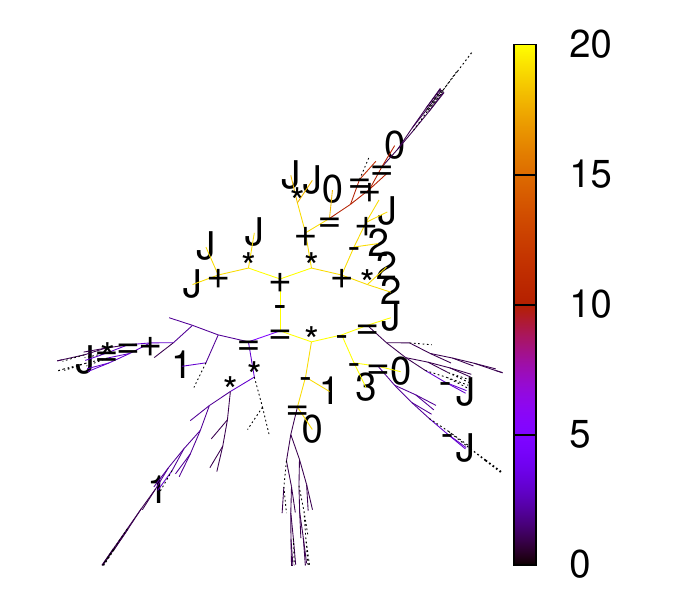}}
\vspace*{-13pt}
\caption{Colour shows location of run time disruption in evolved tree
and number of test cases (1--20) where it impacts the tree's output.
GP trees plotted with output at center of circular lattice
\cite{daida:2005:GPEM}.
Almost all the trees are grey meaning no impact
but only center -10:+10 shown.
=~indicates SRF node, *~multiplication.
Runs 1--5.
Left~+1, right RANDINT.
\label{fig:lattice1}
}
\end{figure*}

\begin{figure*}
\vspace*{-64pt}
\centerline{
\includegraphics{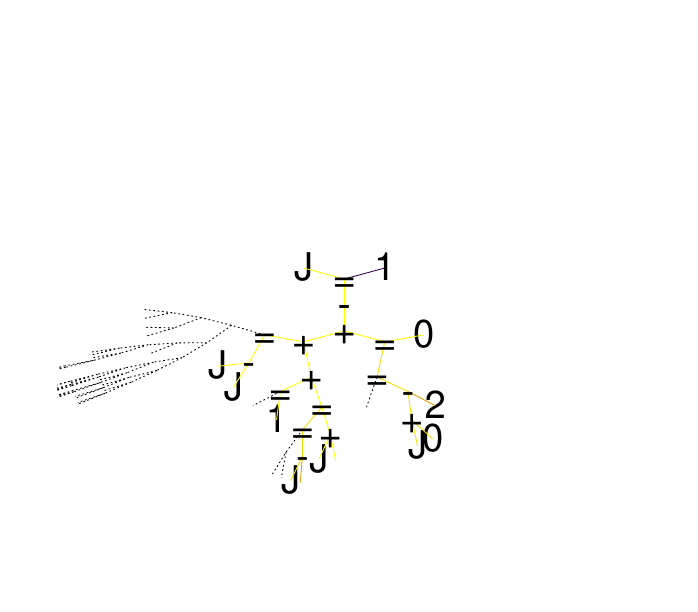}
\includegraphics{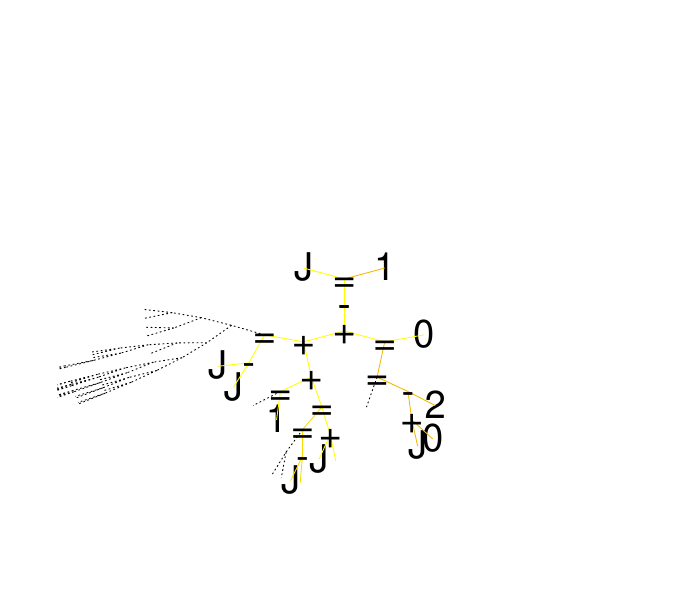}}
\vspace*{\temp}
\centerline{
\includegraphics{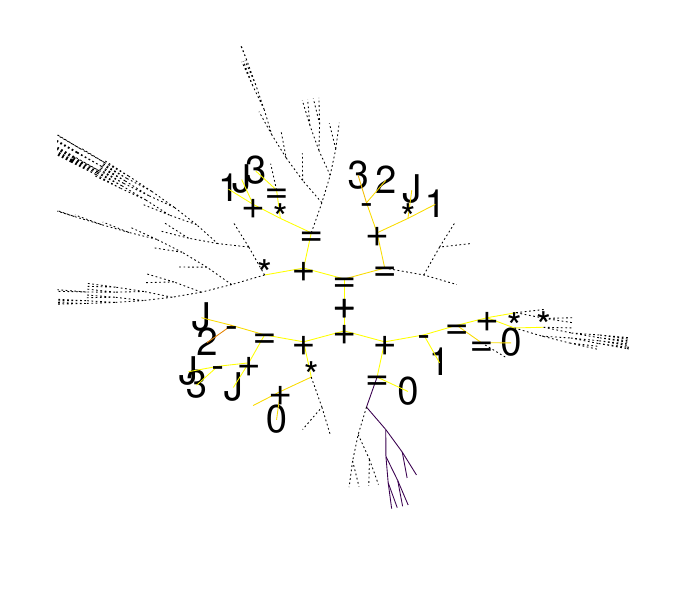}
\includegraphics{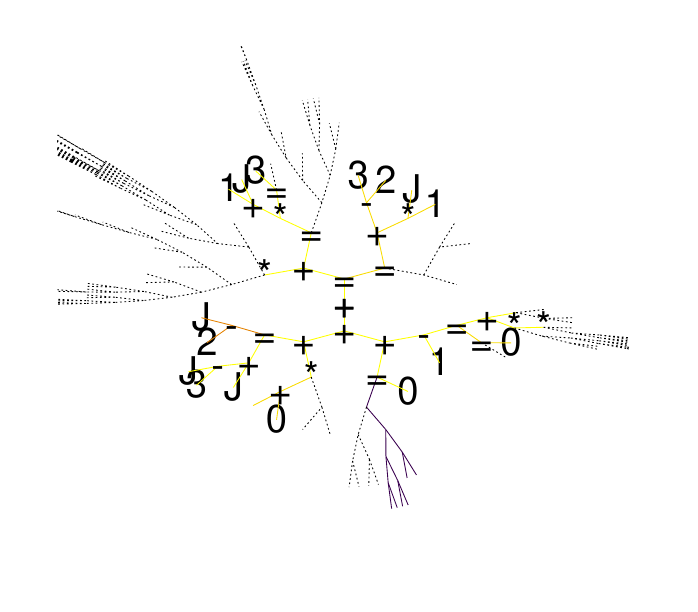}}
\vspace*{\temp}
\centerline{
\includegraphics{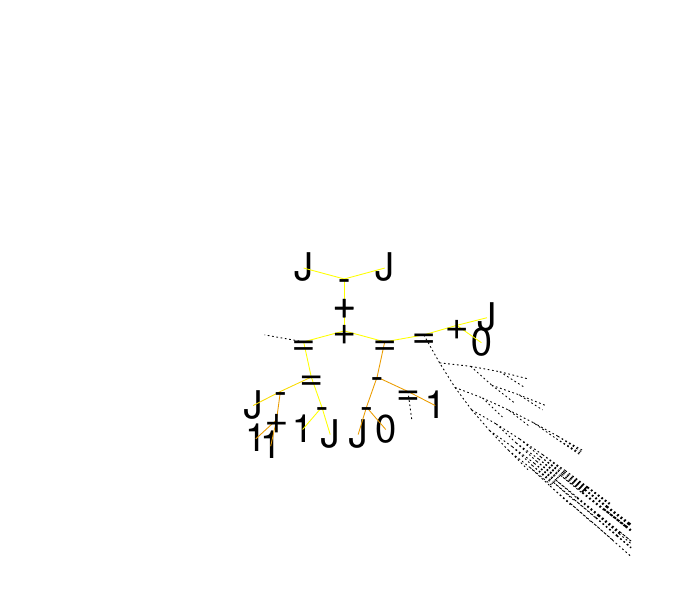}
\includegraphics{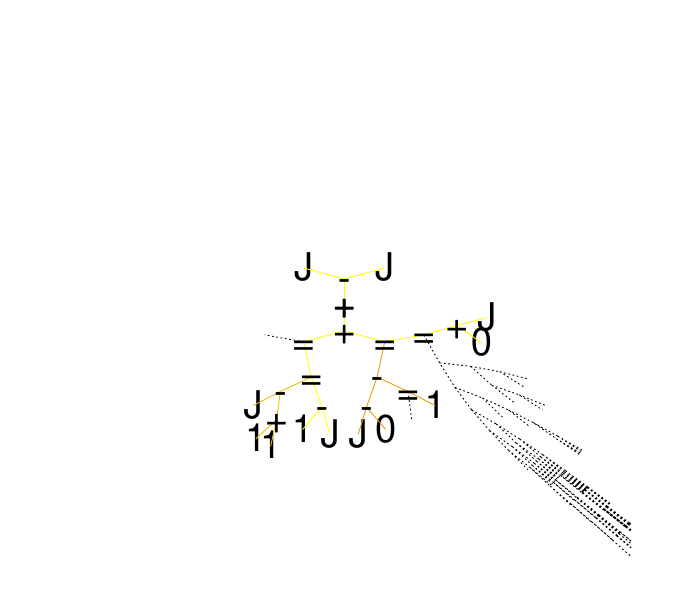}}
\vspace*{\temp}
\centerline{
\includegraphics{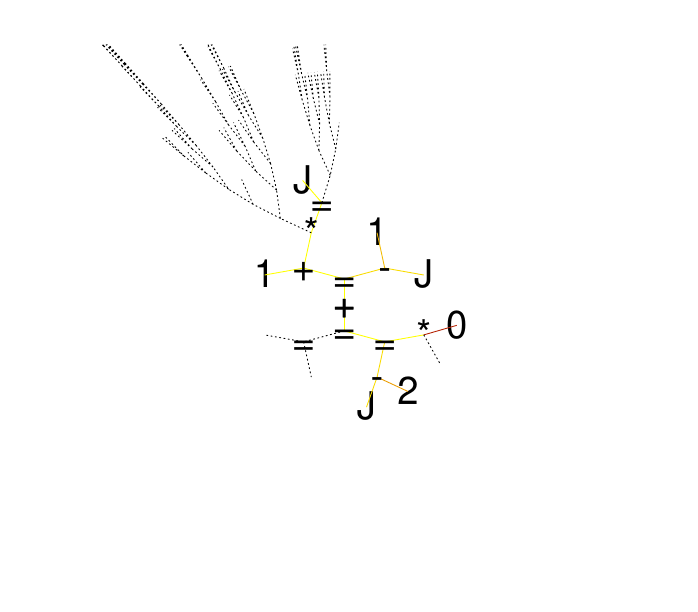}
\includegraphics{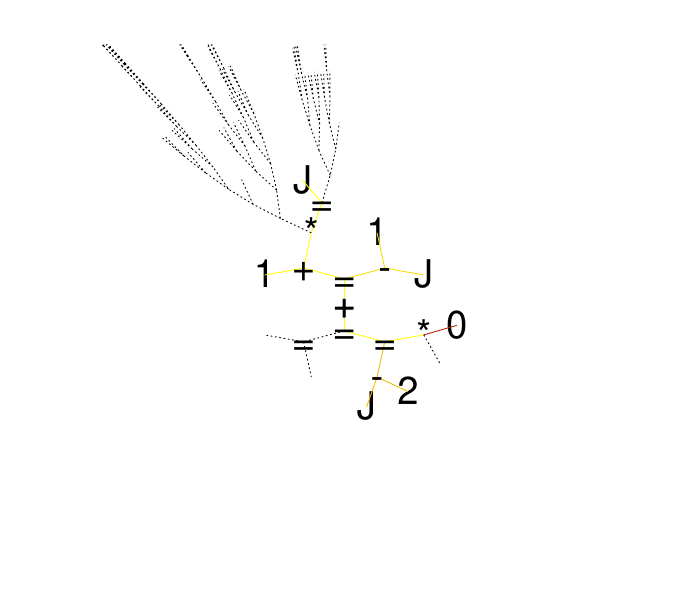}}
\vspace*{\temp}
\centerline{
\includegraphics{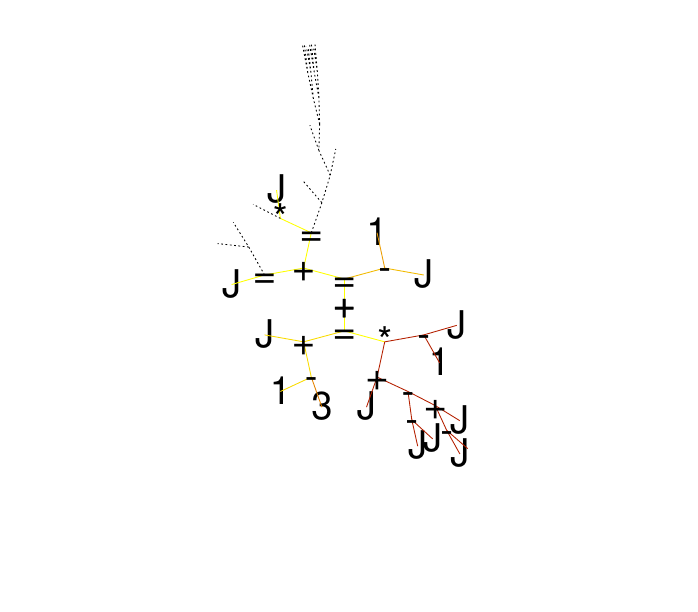}
\includegraphics{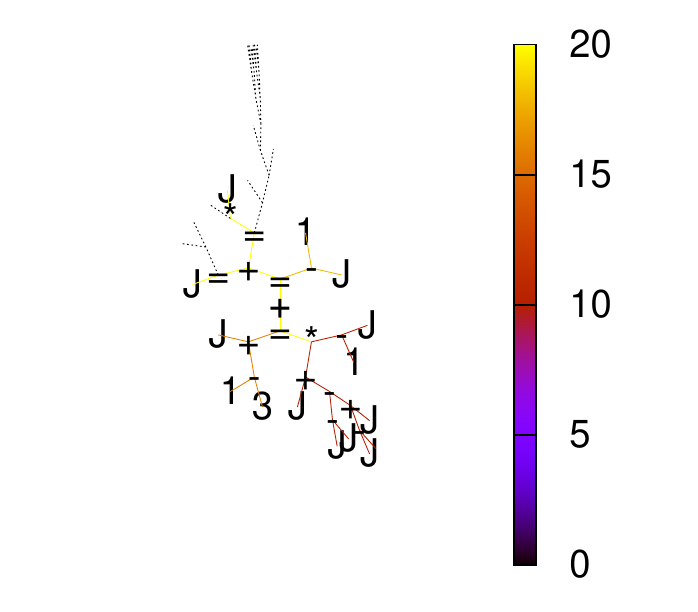}}
\vspace*{-13pt}
\caption{Colour shows location of run time disruption in evolved tree
and number of test cases (1--20) where it impacts the tree's output.
GP trees plotted with output at center of circular lattice
\cite{daida:2005:GPEM}.
Almost all the trees are grey meaning no impact
but only center -10:+10 shown.
=~indicates SRF node, *~multiplication.
Runs 6--10.
Left~+1, right RANDINT.
\label{fig:lattice2}
}
\end{figure*}

As expected, the root node (center) is always disrupted
and so is shown in bright yellow (20 test cases).
The number of test cases disrupted falls monotonically with
distance from the root node.
Dotted gray lines indicates parts of the tree 
where disruption does not reach the root node
on any test case.
For ease of comparison, each plot is shown at the same scale
with parts of the tree outside the square box (-10:$+$10)$^2$
centered on the root node not being plotted.
In half the runs, this box captures all the parts of the tree
where disruption does reach the root node.
Indeed in all runs
all the heavily disrupted parts (bright yellow) 
are plotted.
Only in the third run,
is there code which is disruptive on more than three cases (blue)
which is not plotted
because it lies outside -10:$+$10.

\section{Discussion}
\label{sec:discuss}

At first sight it seems surprising 
in a continuous domain (albeit with exact arithmetic)
that arguably the smallest~(+1) and largest (RANDINT) disruptions
should behave almost identically.
However the two problem dependent
mechanisms for failed disruption propagation
(identified in Section~\ref{sec:mechanisms})
apply to both small and large changes.
That is 
1)~multiplication by zero, gives a zero result,
no matter how small or large the other argument is
and
2)~SRF will return its default value
if its first argument is just out of range 
or if it is widely out of range.

With the small disruption and 32~bit arithmetic,
integer overflow seldom occurs
and if it does,
it makes no difference to
failed disruption propagation.
With the large change,
overflow is more common
but it still make little difference to FDP\@.

We have used various recent efficiency improvements
to Singleton's GPquick
\cite{langdon:2020:small_mem_ga_2pages,
Langdon:2021:GECCO,%
Langdon:2021:GPTP}.
In particular with extended runs~%
\cite{Langdon:2022:ALJ} 
incremental evaluation~\cite{langdon:2021:EuroGP}
again gave substantial speed up
without affecting evolution.

Columns~3 and~4 of Table~\ref{tab:nups_summary}
show again \cite{langdon:1999:aigp3},
but now over an extensive period
(1000 generations rather than \cite{langdon:1999:aigp3}'s 50 or 75),
with two arity functions 
and no size or depth constraints (Table~\ref{gp.details})
that GP evolves
trees whose shape is similar to that of most binary trees~%
\cite{flajlet:1982:ahbt}.

We see highly evolved integer GP programs,
like floating point~\cite{Langdon:2021:LAHS}
and human written code
\cite{langdon:2015:csdc}
\cite{Brownlee:2020:CEC},
are not fragile.
Indeed they are robust not only to genetic changes
like crossover,
but to run time errors,
which they did not encounter during training.

\subsection{Lorenz' butterfly need not trouble software} 

Edward~N. Lorenz (1972)
was uncertain if a single flap of a butterfly's wings in one hemisphere
could cause a dramatic change in the weather the other side of the equator
but argued that the atmosphere is chaotic
and so difficult to predict in the short term
\cite{LorenzButterfly}.

We have argued that deterministic programs are not chaotic.
We have shown often
a single perturbation deep within such software
has no effect.
Whereas the atmosphere's chaos is powered 
by all the Sun's energy falling on the Earth
and contains an unmeasurable number of flapping wings,
deterministic programs (in particular GP)
dissipate information
and so (mostly) give the same result 
even if a ``butterfly'' or other ``bug''
flips some bits deep within them~\cite{danglot:hal-01378523}.

\subsection{Good and bad failed disruption propagation}

The impact of failed disruption propagation is profound.
It is a two edged sword.
One side means crossovers, mutations,
perturbations, radiation, coding errors, etc.,
may only have local impact and their disruption may monotonically
fall to nothing. Making the software robust.
The other edge cuts the tester: Even if an
error, glitch or bug
infects the local state~\cite{Voas:1992:TSE},
if it is
far from the tester's software or hardware probe
the disruption may have faded away before it can be recorded.
Thus rendering the test ineffective
and leaving the error undetected.
However although undetected now,
possibly it may have an effect on a customer later.

\subsection{Better Evolutionary Computation?}

In tree GP~\cite{koza:book},
almost all crossovers or mutations occur near leafs
far from the root node.
In deep trees,
their impact is often lost before 
it reaches the program's output (the root node).
In artificial systems we are free to choose where to place
mutations and where to cut and slice in recombination.
So we could opt to place such disruption close to the root node.
However, if we are to evolve complex programs with many many features,
they will have to be large.

Evolving a monolithic one or two dimensional structure
with all information channeled via a single output node
risks the program being so deep that it is impossible to measure the fitness
of most genetic updates,
or, if we move the crossover locations to be by the output node, 
we have the problem of
carrying a large dead weight of code which cannot adapt
beneath a tiny living evolving surface near the route node.
Therefore instead we may want to adopt a porous open sponge like 
high dimensional structure,
with a large surface area,
where much of the program 
(and hence most of the genetic locations)
is close to the program's environment
\cite{langdon:2021:sigevolution}.

\section{Conclusions}
\label{sec:conclude}

We have measured 
software engineering's
failed disruption propagation (FDP)~\cite{Petke:2021:FSE-IVR}
in genetic programming
and find as predicted by information theory
in integer functions
it is very common.
On average in our deep trees 99.7\% of 
large run time disruptions fail to propagate to the root node
and so have no impact on fitness.

We see failed disruption propagation scaling at between 
$e^{-{\rm depth}/3}$ and~$e^{-{\rm depth}/5}$,
meaning the chance of detecting disruption 
(be it induced by crossover, mutation, cosmic ray or indeed software bug)
falls significantly within 3 to~5 levels.
Indeed, with every extra level of nesting
the effectiveness of
optimal test oracle placement
or fitness measurement
falls by between 18\% and~28\%.

We see little difference between the smallest possible disruption
(Section~\ref{sec:+1})
and total runtime randomization
(Section~\ref{sec:RANDINT}),
suggesting Danglot et al.'s~\cite{danglot:hal-01378523}
correctness attraction
will hold more widely than their +1 disruption.
Indeed these experiments with integer functions,
support the view that software is not fragile~%
\cite{langdon:2015:csdc}.

The average depth,
rather than size,
is critical.
With nesting deeper than 5--7 levels 
it becomes impossible to see the effect
of most individual crossovers or mutations
and the fitness landscape becomes increasingly flat
and evolution harder.
This suggests either the need to limit the depth of
crossover and mutation or the need to move fitness testing
from the root node to
closer to the genetic changes~%
\cite{langdon:2021:sigevolution}.

\subsection*{Acknowledgments}

This work was inspired by conversations
at Dagstuhl 
\href{https://www.dagstuhl.de/en/program/calendar/semhp/?semnr=18052}
{Seminar 18052} on
Genetic Improvement of Software
\cite{gi_dagstuhl_2018}.
I am grateful for the assistance of anonymous reviewers.
Supported by the Meta OOPS project.

\noindent
GPQuick code is
available in
\href{http://www.cs.ucl.ac.uk/staff/W.Langdon/ftp/gp-code/GPinc.tar.gz}
     {http:/\allowbreak{}/\allowbreak{}www.cs.ucl.ac.uk/\allowbreak{}staff/\allowbreak{}W.Langdon/\allowbreak{}ftp/\allowbreak{}gp-code/\allowbreak{}GPinc.tar.gz}

\bibliographystyle{ACM-Reference-Format}
\bibliography{/tmp/references,/tmp/gp-bibliography}

\end{document}

%% file: graph/nups_summary2.tex
 86035 &  663 & 735 & 160 &   20  &  0.114  &-0.31 
&  0.092  &-0.31 
\\
  4347 &  160 & 165 &  36 &   10  &  1.449  &-0.30 
&  1.449  &-0.33 
\\
 23289 &  220 & 383 &  83 &  184  &  3.010  &-0.27 
&  3.053  &-0.27 
\\
131159 &  449 & 908 & 197 &  130  &  0.127  &-0.28 
&  0.121  &-0.29 
\\
 77479 &  454 & 698 & 152 &  632  &  0.253  &-0.20 
&  0.256  &-0.20 
\\
 51697 &  626 & 570 & 124 &    0  &  0.056  &-0.27 
&  0.056  &-0.27 
\\
   771 &   33 &  64 &  14 &    0  &  7.523  &-0.21 
&  7.523  &-0.22 
\\
 35727 &  425 & 474 & 103 &    0  &  0.073  &-0.30 
&  0.073  &-0.30 
\\
 53305 &  485 & 579 & 126 &    0  &  0.032  &-0.33 
&  0.032  &-0.33 
\\
 23377 &  360 & 383 &  83 &    0  &  0.137  &-0.26 
&  0.137  &-0.26 